# Deep Learning Framework for Early Detection of Pancreatic Cancer Using Multi-Modal Medical Imaging Analysis

[1]Dennis Slobodzian, [*1]Amir Kordijazi

[*]amir.kordijazi@maine.edu

[1]Department of Engineering, University of Southern Maine, P.O. Box 9300, Gorham, ME 04038, USA

## Abstract

Pancreatic ductal adenocarcinoma (PDAC) remains one of the most lethal forms of cancer, with a five-year survival rate below 10% primarily due to late detection [1]. This research develops and validates a deep learning framework for early PDAC detection through analysis of dual-modality imaging: autofluorescence and Second Harmonic Generation (SHG). We analyzed 40 unique patient samples to create a specialized neural network capable of distinguishing between normal, fibrotic, and cancerous tissue. Our methodology evaluated six distinct deep learning architectures, comparing traditional Convolutional Neural Networks (CNNs) with modern Vision Transformers (ViTs). Through systematic experimentation, we identified and overcame significant challenges in medical image analysis, including limited dataset size and class imbalance. The final optimized framework, based on a modified ResNet architecture with frozen pre-trained layers and class-weighted training, achieved over 90% accuracy in cancer detection. This represents a significant improvement over current manual analysis methods and demonstrates potential for clinical deployment. This work establishes a robust pipeline for automated PDAC detection that can augment pathologists' capabilities while providing a foundation for future expansion to other cancer types. The developed methodology also offers valuable insights for applying deep learning to limited-size medical imaging datasets, a common challenge in clinical applications.

## 1 Introduction

Pancreatic ductal adenocarcinoma (PDAC) represents ~90% of all pancreatic cancer cases and continues to challenge oncologists, with a five-year survival rate remaining below 10% despite concerted research efforts [2]. This persistently grim prognosis is driven largely by late-stage detection; by the time symptoms emerge, nearly 80% of patients have locally advanced or metastatic disease that severely limits therapeutic options [1].

Current diagnostic workflows—relying on CT, MRI, ultrasound, PET imaging, and manual histopathology—struggle with early-stage and subtle lesion detection, especially in differentiating benign fibrosis from emerging tumors. Computed tomography, the primary imaging modality, fails to detect approximately 40% of tumors smaller than 2 cm, with human error from interobserver variability compounding the challenge [3]. Retrospective studies show that these limitations often result in delayed diagnosis and reduced treatment efficacy.

Recent progress in artificial intelligence, especially through deep learning (DL) and radiomics, has shown promise in improving detection and characterization of pancreatic lesions on CT and MRI [3–16]. Despite impressive retrospective results, many DL models are trained on single-center datasets, raising concerns about overfitting, annotation bias, and lack of standardization. Broader adoption is hampered by limited data sharing, poor external validation, and obstacles in integrating AI into clinical workflows.

Emerging optical modalities like autofluorescence and Second Harmonic Generation (SHG) offer high-resolution insights into tissue microstructure and collagen organization—key indicators of early tumorigenesis [10,17–19]. Prior work has applied SHG to identify fibrotic changes associated with cancer in diverse tissues through various types of analysis such as collagen fiber alignment and texture [17–22]. To mitigate concerns about superficial optical imaging, several research groups are designing optical endoscopes suitable for SHG and autofluorescence imaging [23–25]. Here we initiate a dual-imaging based deep learning model on PDAC pathology slides to provide a classification algorithm that may ultimately be combined with endoscopic dual modality imaging to aid in earlier diagnosis of PDAC to improve patient care. [26].

This research aims to develop and validate an automated system for early PDAC detection. First, we focus on developing a specialized neural network architecture optimized for dual-modality opticalimaging analysis of 40 pancreatic biopsies supplied by the Maine Medical Center BioBank (MMC BB) with a target goal of at least 90% classification accuracy on a validation dataset[27]. Second, we have a dataset development and validation system, processing both the autofluorescence and SHG images.

## 2 Methodology

### 2.1 Dataset Acquisition and Characteristics

This study utilized a dataset obtained from the Maine Medical Center BioBank (MMC BB), consisting of forty H&E stained slides of pancreatic tissue from unique patients. Each slide was scanned using an Aperio2 slide scanner and annotated by a pathologist, who identified regions of cancer, fibrosis, and normal tissue (Figure 1 (a)). Twenty slides were annotated as only normal tissue, while the other twenty contained cancerous regions. Of these cancer-containing slides, fifteen also had areas annotated as fibrosis, and thirteen contained regions of normal adjacent tissue. **Figure 1** shows examples of the three primary tissue categories i.e., normal, and cancerous tissues.

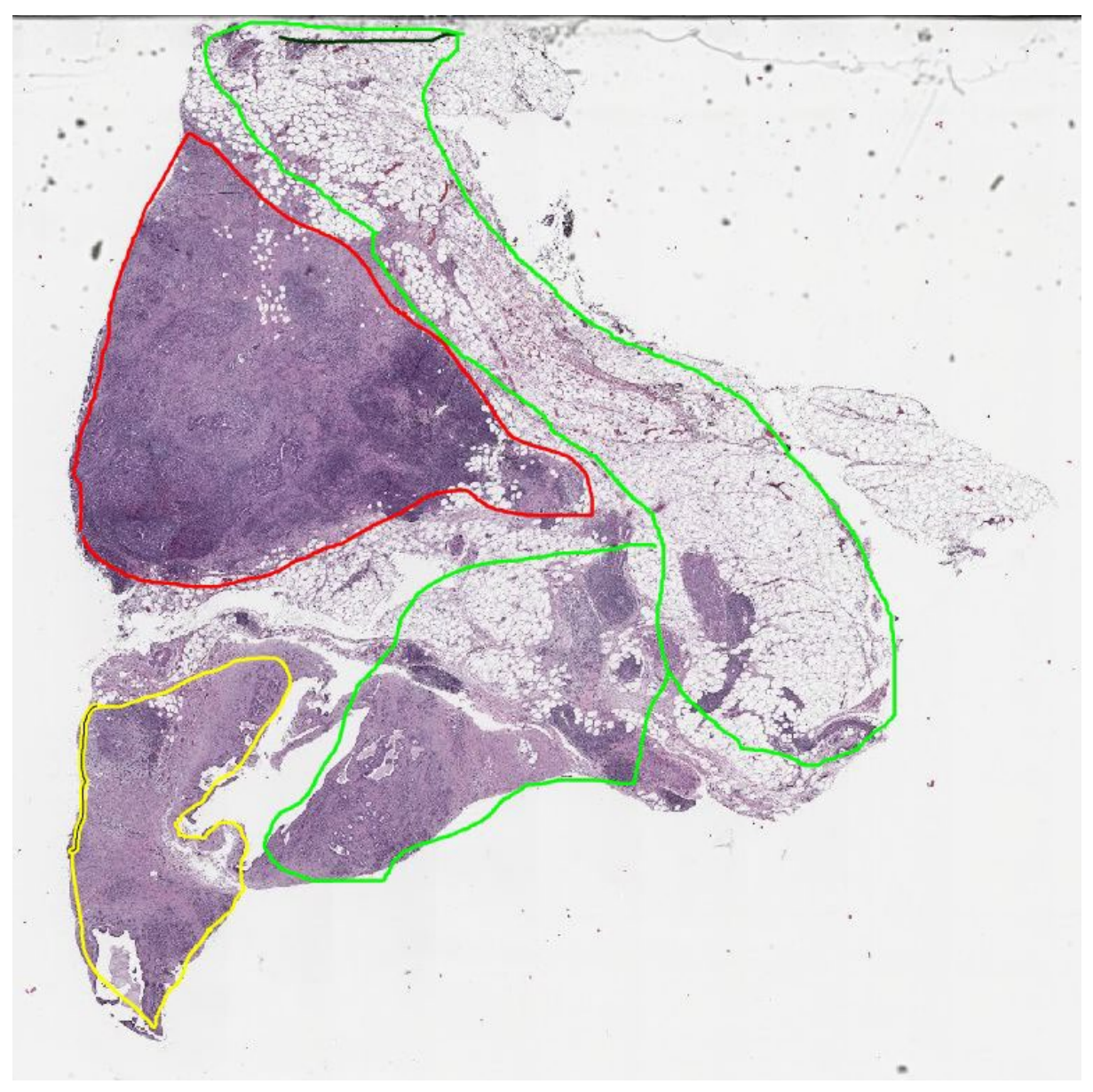

(a)

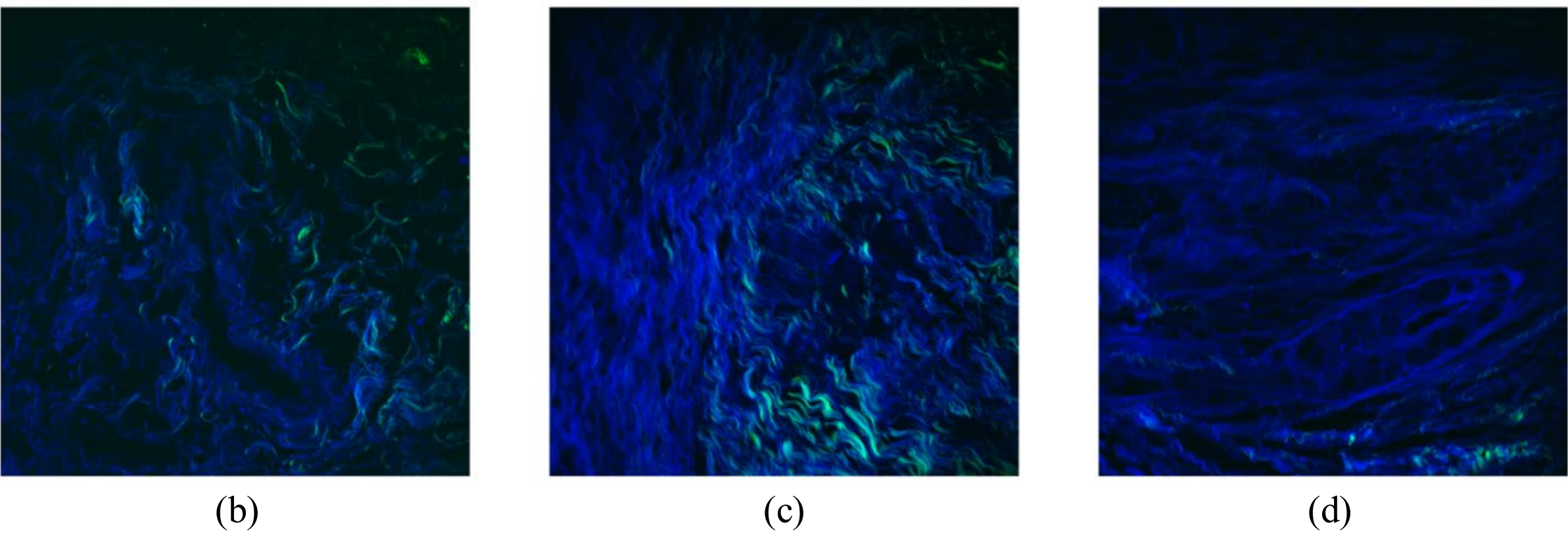

**Figure 1**. (a) An H&E stained slide of pancreatic tissue annotated by a pathologist, where red indicates cancer, yellow indicates fibrosis, and green indicates normal adjacent tissues. Representative images showing the three primary tissue categories: (b) Normal pancreatic tissue showing regular ductal structure, (c) Fibrotic tissue displaying characteristic stromal patterns, and (d) Cancerous tissue exhibiting PDAC morphology. Images captured using dual-modality imaging combining autofluorescence and SHG signals.

Eight ductal structures were imaged within each tissue category, resulting in a final dataset of 36 normal tissue images, 102 cancer images, and 101 fibrotic tissue images. Each region was imaged using two complementary imaging techniques: autofluorescence imaging to capture metabolic indicators through natural tissue fluorescence, and Second Harmonic Generation (SHG) imaging to reveal structural features through collagen visualization. **Figure 2** demonstrate the comparison of the imaging modalities i.e., Autofluorescence, and SHG.

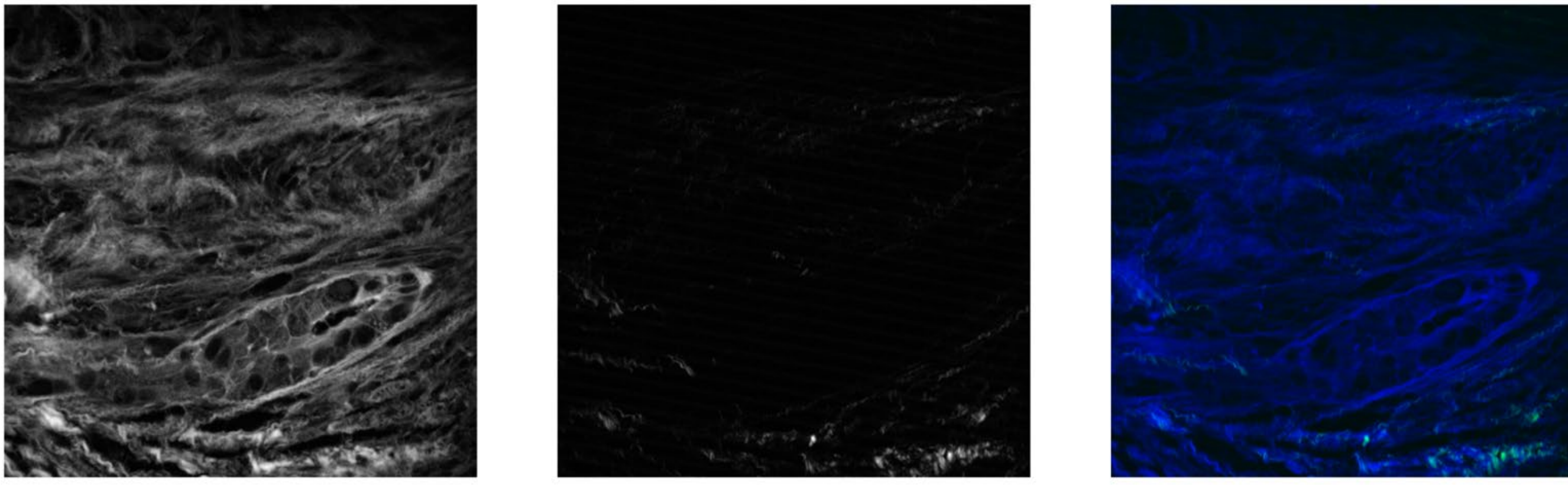

**Figure 2**. Comparison of imaging modalities showing different aspects of tissue architecture: (Left) Autofluorescence signal highlighting cellular metabolic activity and proteins such as collagen crosslinks, and elastin, (Center) SHG imaging revealing collagen structure, (Right) Combined visualization integrating both modalities.

### 2.2 Image Processing Pipeline

The development of a preprocessing pipeline proved crucial for maintaining data consistency and quality. Raw images undergo a series of preprocessing steps to ensure standardization across the dataset. Initial normalization scales both autofluorescence and SHG channels to a standardized [0, 255] range, preserving relative intensities while enabling consistent processing. The channel combination process applies specific scaling factors (1.3 for SHG in the green channel, 1.0 for autofluorescence in the blue channel) to optimize visual contrast and feature distinction.

The resulting dual-channel images have further processing through a data augmentation pipeline designed to preserve biological relevance while increasing dataset diversity. This includes carefully controlled geometric transformations, limited to ±30 degrees rotation to not greatly modify the structure of the samples. These parameters were chosen to reflect natural variations in tissue preparation and imaging conditions without introducing artifacts that could mislead the learning process.

### 2.3 Deep Leaning Architecture

Our experimental approach began with an evaluation of deep learning architectures to determine the most effective foundation for pancreatic tissue analysis. We investigated six distinct architectures, divided into two main categories: i.e., CNN-based and Vision Transformers architecture as listed in **Table 1**.

**Table 1**. Deep learning algorithms used for image analysis

| Deep Learning Architectures | Note |
|---|---|
| CNN-based architectures | |
| – ResNet-50 | Built on residual connections to enable deeper network training |
| – DenseNet-121 | Employing dense connectivity for efficient feature reuse |
| – EfficientNet-B0 | Using compound scaling for balanced network growth |
| Vision Transformer architectures | |
| – DINOv2 | Leveraging self-supervised learning capabilities |
| – Swin Transformer | Using hierarchical feature representation |
| – Standard ViT | Implementing pure attention-based processing |

Each architecture underwent initial training using transfer learning from ImageNet pretrained weights, maintaining identical classifier heads to ensure fair comparison. This phase showed fundamental challenges in applying deep learning to our limited medical imaging dataset. While all architectures achieved high training accuracy, their validation performance plateaued around 13%, indicating severe overfitting. The Vision Transformers particularly struggled with the limited data, likely due to their larger parameter counts and lack of inductive biases present in CNNs [28]. **Figure 3** shows training and validation accuracy of six model architectures. These findings pushed us to explore techniques specifically designed for limited data scenarios.

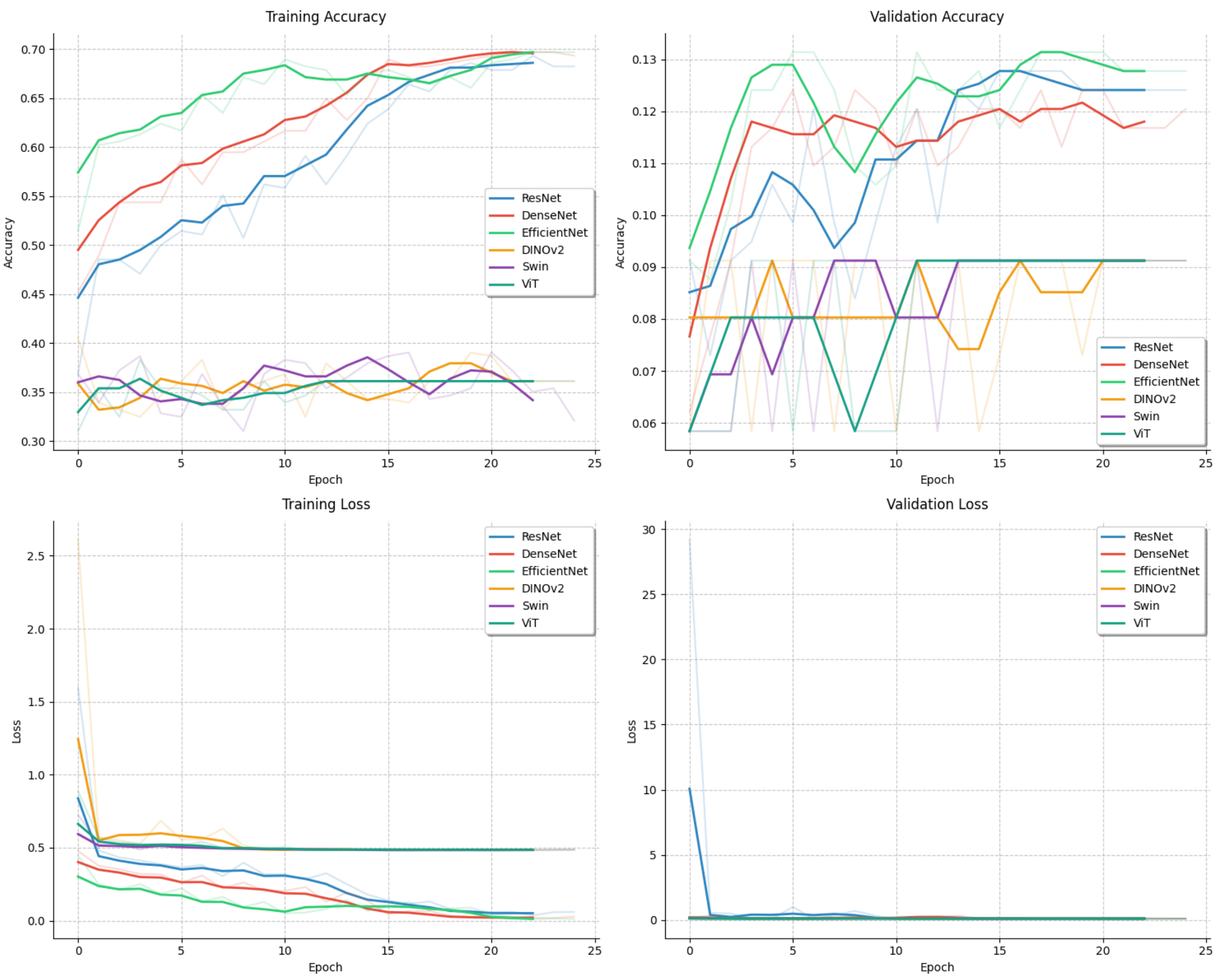

**Figure 3**: Initial comparison of model architectures showing training and validation accuracy. CNN-based models demonstrated better convergence, but all models showed significant overfitting with validation accuracies plateauing around 13%. The clear gap between training and validation performance indicated the need for a more robust training strategy.

### 2.4 Feature Extraction through Frozen Backbones

The clear overfitting in our initial experiments led us to modify our approach fundamentally. Rather than allowing the entire network to learn from our limited dataset, we froze the pretrained backbone layers, effectively treating the networks as feature extractors. This approach leverages the robust feature detection capabilities learned from ImageNet while limiting the learn-able parameters to just the classification layers. **Figure 4** shows training and validation accuracy of six model architectures after fine-tuning.

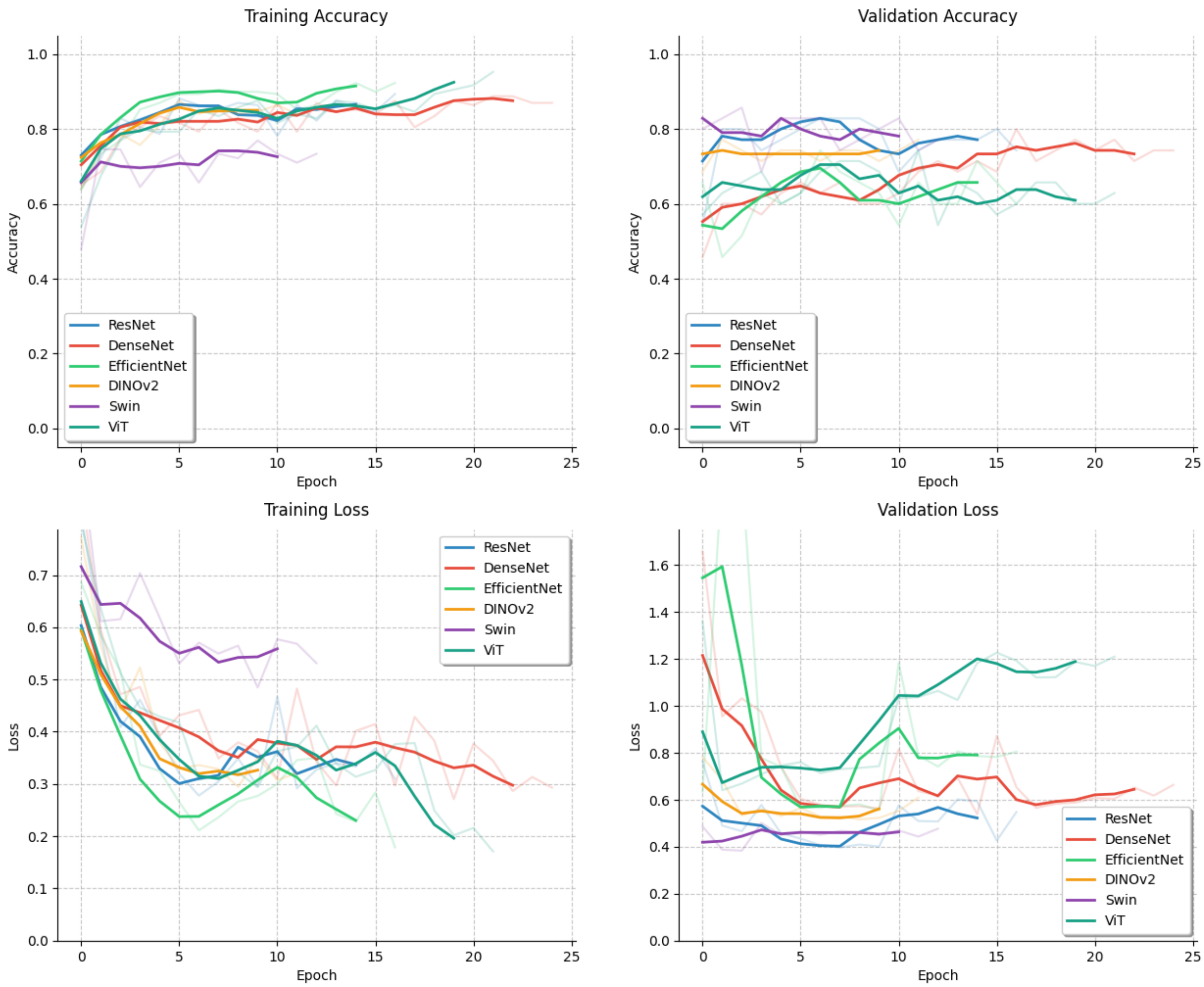

**Figure 4**. Performance comparison with frozen backbones. This approach significantly improved validation accuracy to 60-80% while reducing the gap between training and validation performance, indicating better generalization.

The results of this modification proved transformative. Validation accuracies increased dramatically from 13% to 60-80%, while the gap between training and validation performance decreased. Among all architectures, ResNet emerged as the most stable platform, consistently delivering reliable performance across multiple training runs. This improvement confirmed our idea that ImageNet-learned features could transfer effectively to medical imaging tasks when properly constrained.

### 2.5 Addressing Class Imbalance

The use of our frozen backbone approach revealed another subtle but critical challenge: our model showed bias in its predictions due to the uneven distribution of our dataset. With normal tissue samples significantly underrepresented compared to cancer and fibrosis samples, we implemented a weighted training strategy to balance the learning process as demonstrated in **Figure 5**.

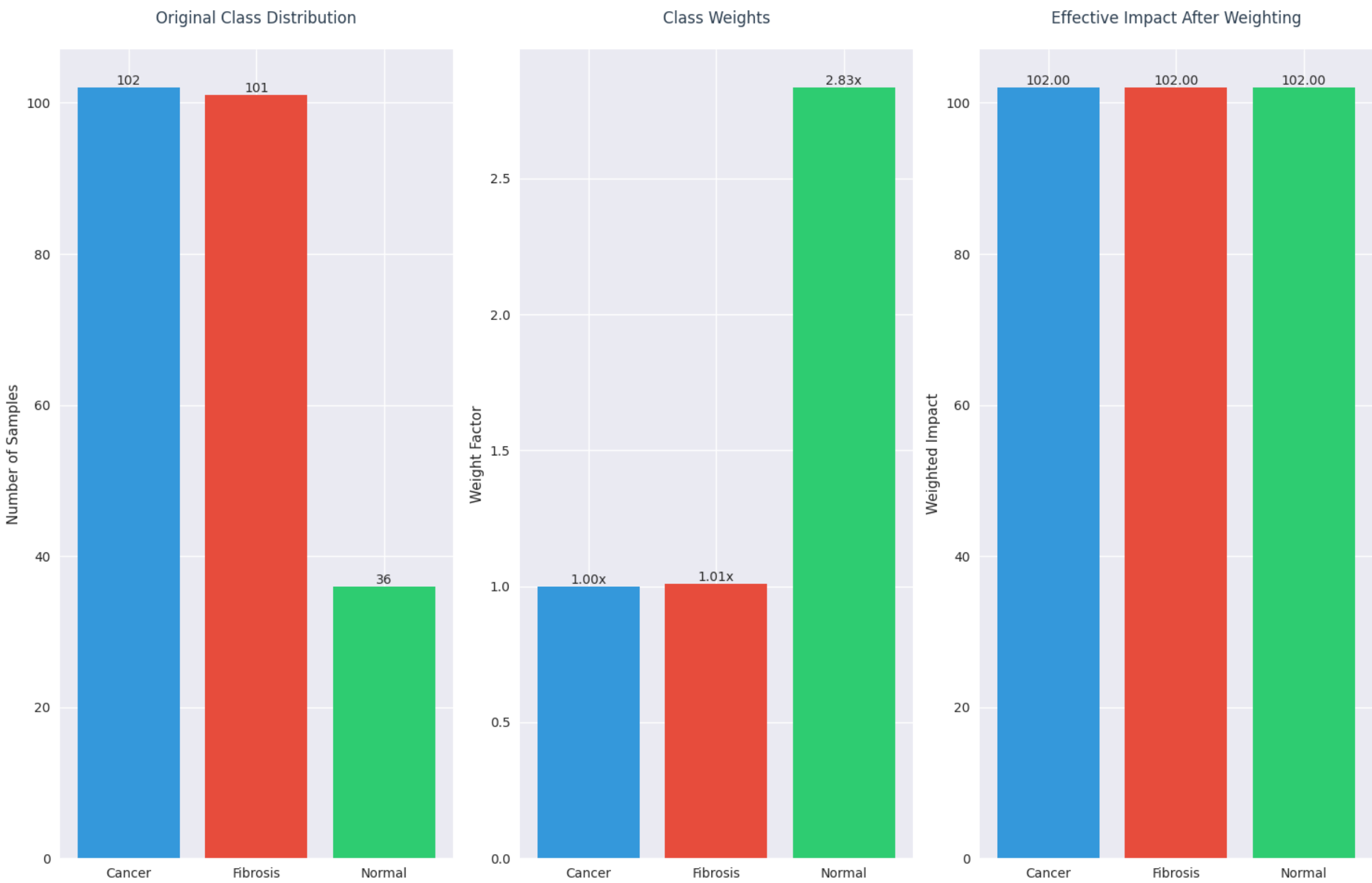

Figure 5: Dataset balancing strategy showing original class distribution, applied class weights, and effective balanced impact during training. Normal tissue samples were weighted 2.83x higher to compensate for their underrepresentation.

Our weighting idea assigned importance factors inversely proportional to class frequency, with normal tissue samples weighted 2.83 times higher than cancer samples. This proved crucial in achieving balanced performance across all tissue types while maintaining the model's overall accuracy. The weighted approach particularly improved the detection of normal tissue samples without compromising the model's performance on majority classes.

### 2.6 Robust Validation through K-Fold Implementation

The final phase of our experimental progression focused on ensuring the reliability of our results given our limited dataset size. We implemented a 5-fold cross-validation strategy, evaluating our model's performance across different data splits [29].

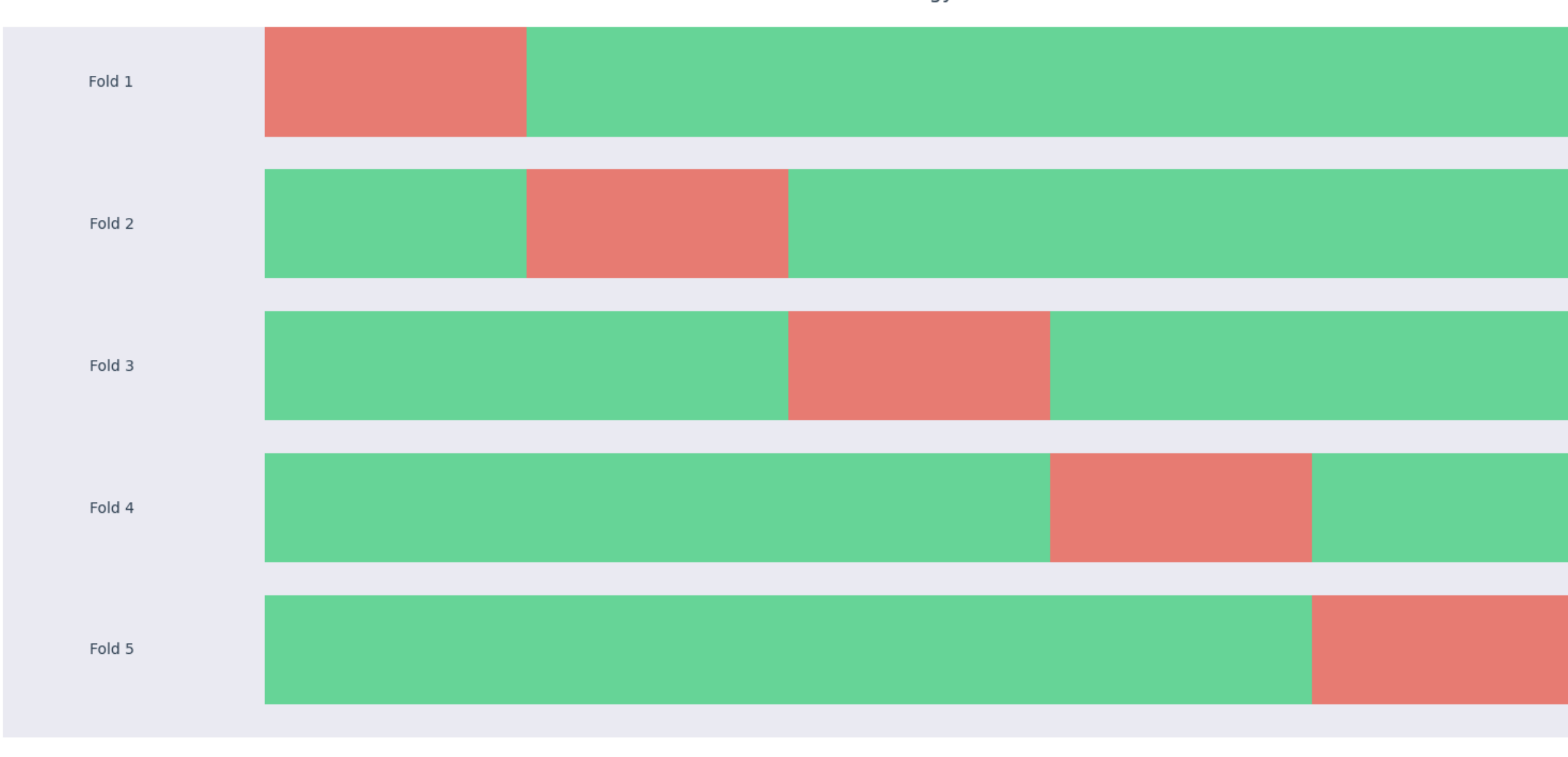

Figure 6: K-fold cross-validation strategy illustrating how the dataset was divided into five portions, with each portion serving as validation data once. This approach provided more reliable performance estimates and helped identify potential biases in our training process.

The k-fold validation process reinforced our confidence in the model's performance, demonstrating consistent results across different data splits and initializations. This comprehensive validation approach provided strong evidence of our model's generalization capabilities. Each fold's results remained stable, suggesting that our previous optimizations had indeed created a robust framework for pancreatic tissue classification.

This methodical progression from basic transfer learning to a fully optimized pipeline demonstrates the importance of systematically addressing each challenge in medical image analysis. The framework combines frozen backbones for feature extraction, weighted training for balanced learning, and comprehensive validation for reliability assessment, creating a robust solution for pancreatic tissue analysis.

## 3 Results and Discussion

### 3.1 ResNet Architecture Analysis

After establishing our balanced training approach and k-fold validation strategy, we conducted a systematic comparison of ResNet architectures to determine the best network depth for pancreatic tissue classification. This compared ResNet-18, ResNet-34, and ResNet-50, evaluating both performance metrics and computational efficiency as shown in **Figure 7**.

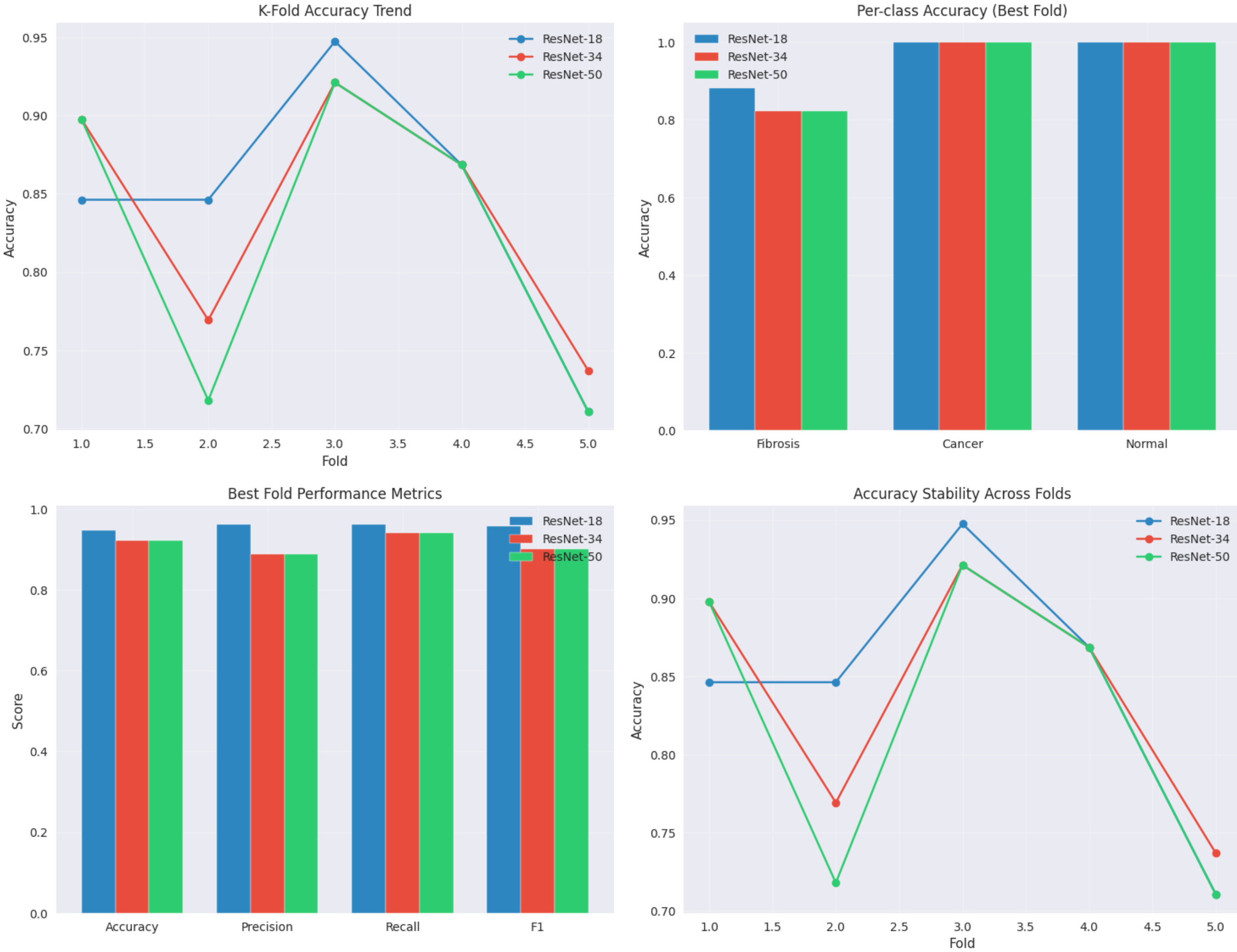

**Figure 7**: Performance analysis of ResNet architectures showing (a) k-fold accuracy trends, (b) per-class accuracy for best performing fold, (c) overall performance metrics, and (d) accuracy stability across folds.

**3.1.1 Model Stability Analysis:** The k-fold accuracy trends revealed interesting insights about model stability. ResNet-18, despite being the simplest architecture, maintained the most consistent performance across folds. In contrast, both ResNet-34 and ResNet-50 showed greater performance variance, particularly in later folds where accuracy dropped significantly. This pattern suggests that deeper architectures, while more powerful, may be more susceptible to overfitting on our dataset.

**3.1.2 Per-class Performance:** Analysis of per-class accuracy revealed that ResNet-18 achieved more balanced performance across tissue types. For the best-performing fold, ResNet-18 showed strong consistency, achieving approximately 0.88 accuracy for fibrosis, 1.0 for cancer, and 1.0 for normal tissue during validation. The deeper architectures, while capable of similar peak performance, showed greater differences between classes, having potential issues with class bias.

**3.1.3 Performance Metrics:** Examination of key performance metrics revealed that ResNet-18 consistently outperformed its deeper counterparts in terms of precision, recall, and F1 score (**Table 2**).

**Table 2**. Performance metrics for ResNet-18

| Performance Metrics | Value |
|---|---|
| Overall accuracy (in its best fold) | 0.947 |
| Macro-averaged precision | 0.961 |
| Macro-averaged recall | 0.961 |
| Macro-averaged F1 score | 0.958 |

These results challenge the common assumption that deeper networks necessarily perform better. In our specific case, the simpler ResNet-18 architecture proved more effective, likely due to several factors:

1. Dataset Size: Our relatively small dataset (239 samples) may not provide sufficient data for deeper networks to learn meaningful features without overfitting.

2. Feature Complexity: The visual features distinguishing pancreatic tissue types may be sufficiently captured by the simpler architecture, making additional depth unnecessary.

3. Training Stability: The reduced parameter count in ResNet-18 resulted in more stable training dynamics and better generalization.

Based on these findings, we selected ResNet-18 as our primary architecture, focusing subsequent optimizations on this model. This choice emphasizes the importance of matching model complexity to dataset characteristics rather than defaulting to deeper architectures. The superior performance of ResNet-18 also suggests that pancreatic tissue classification may rely more on fundamental structural features that can be effectively captured by simpler network architectures.

### 3.2 Balanced Training Implementation

Following the selection of ResNet-18 as our primary architecture, we implemented a balanced training strategy to address the inherent class imbalance in our dataset. This approach weighted the loss function inversely proportional to class frequencies, with particular emphasis on normal tissue samples which were underrepresented in the original dataset distribution.

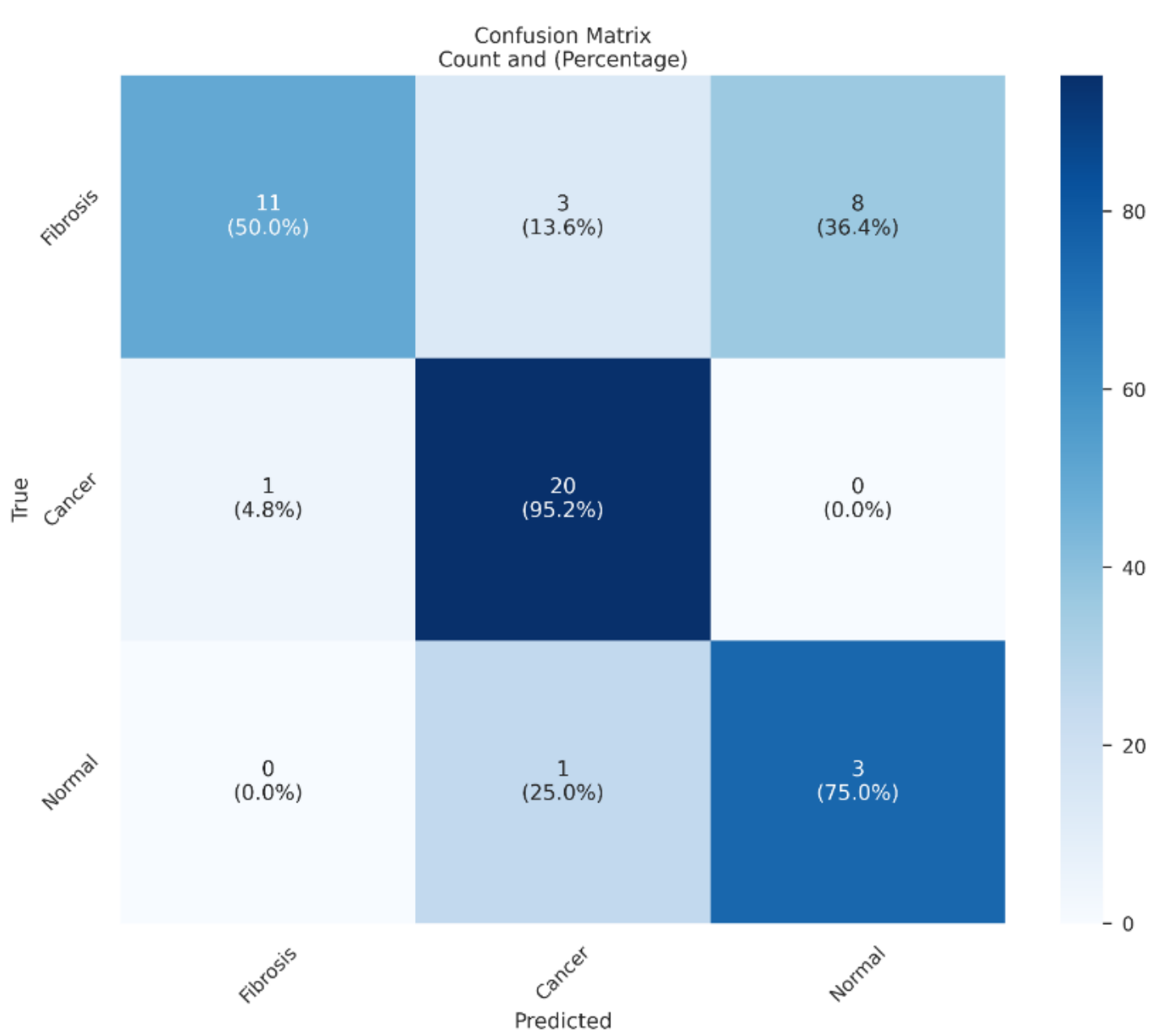

**Figure 8**. Confusion matrix showing classification outcomes after balanced training implementation. While cancer detection maintains strong performance (95.2% accuracy), the results highlight persistent challenges in normal tissue classification.

The confusion matrix (Figure 8) results revealed both strengths and persistent challenges in our approach. Cancer detection maintained impressive accuracy at 95.2% (20 out of 21 samples), demonstrating robust performance in identifying malignant tissue patterns. However, normal tissue classification continued to present challenges, with significant confusion between tissue types leading to reduced precision (0.273).

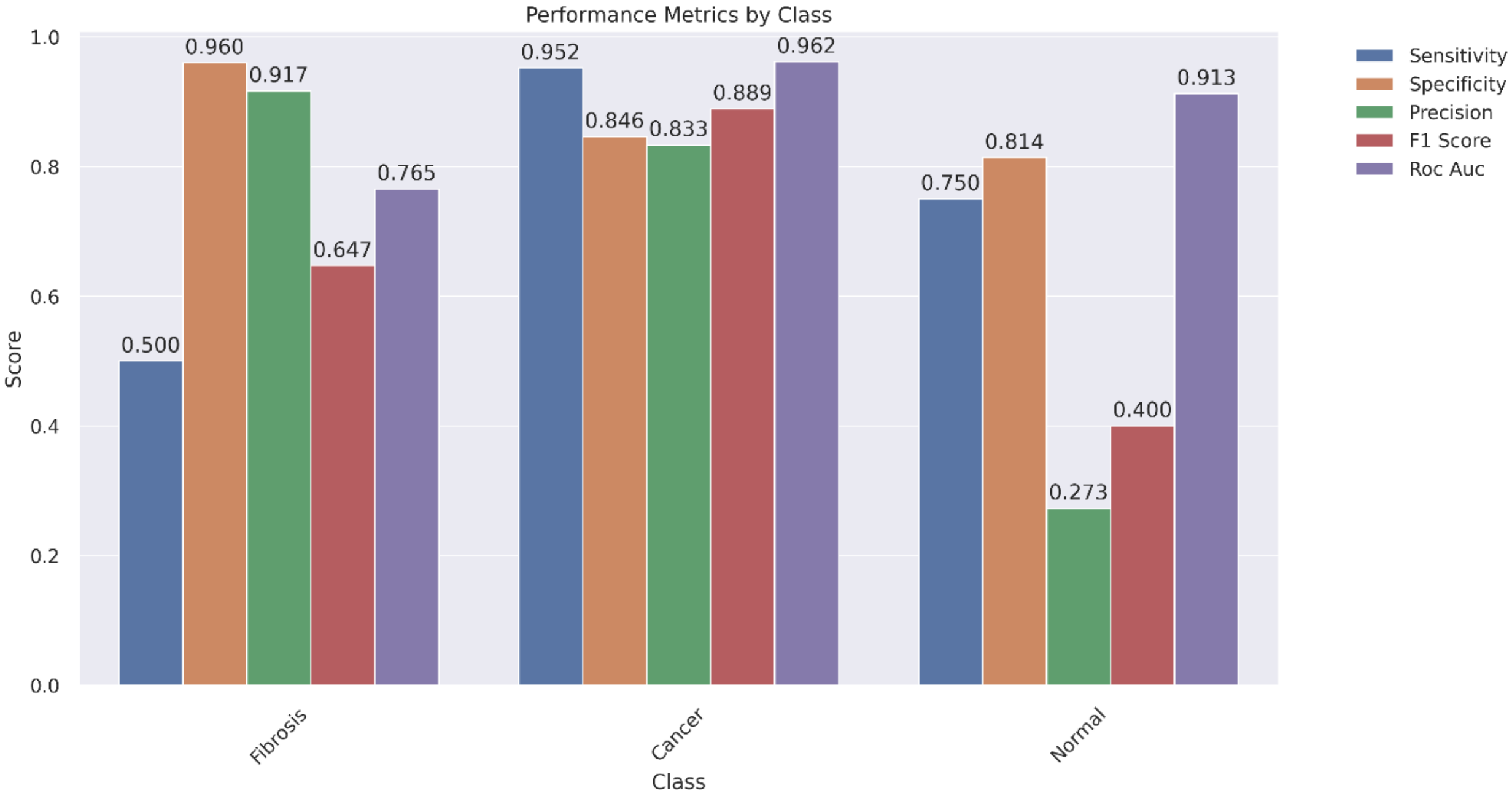

**Figure 9**. Performance metrics across tissue types showing varying levels of success despite balanced training. Cancer detection maintains strong metrics while normal tissue classification shows room for improvement.

Detailed analysis of performance metrics shown in **Figure 9**, revealed distinct patterns across tissue types. Cancer identification remained highly reliable, while fibrosis detection showed moderate success, and Normal tissue classification presented ongoing challenges.

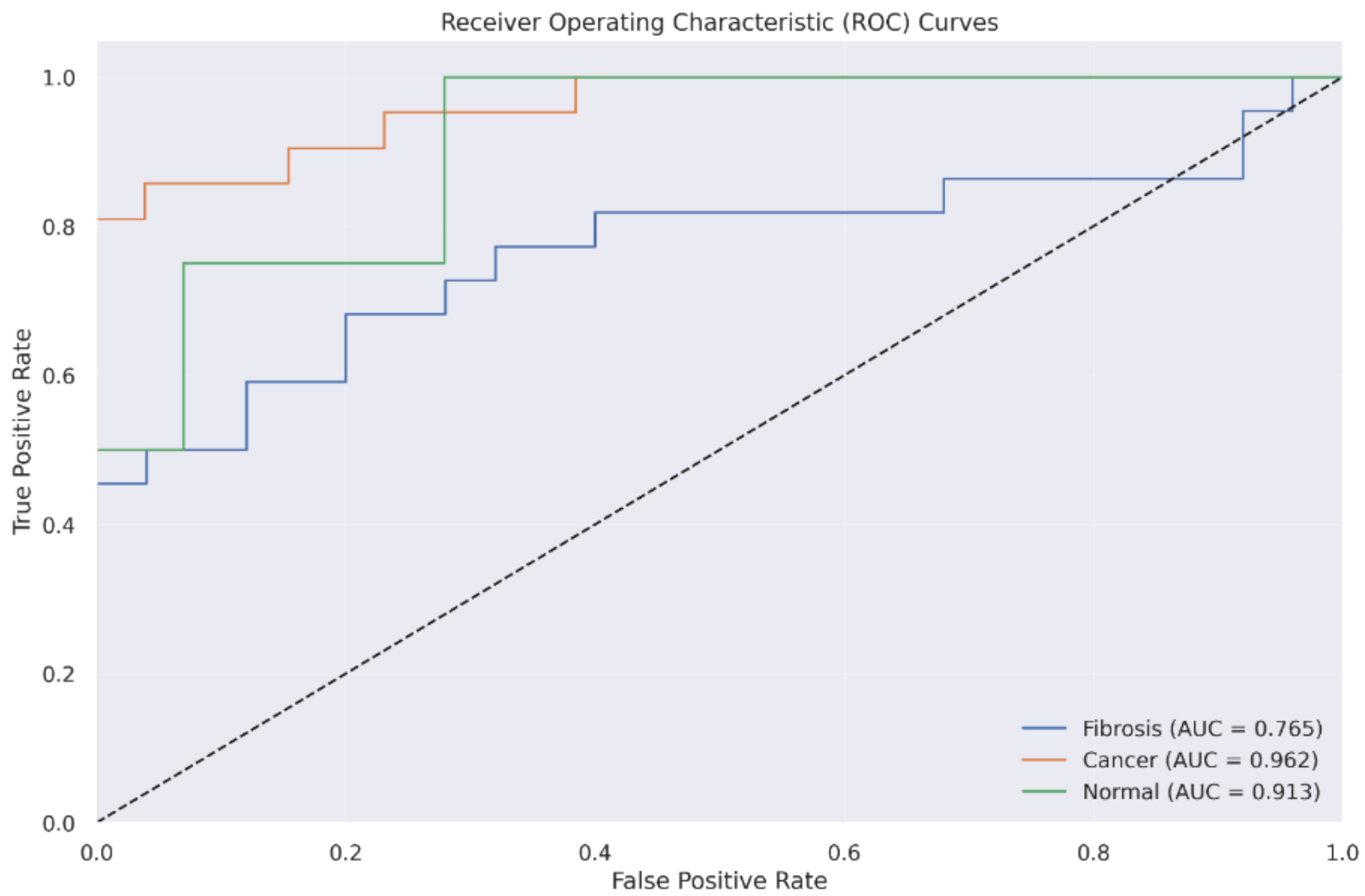

Figure 10: ROC curves demonstrating classification performance characteristics. While cancer detection shows excellent discrimination (AUC = 0.962), fibrosis and normal tissue classification exhibits a less accurate performance.

The ROC curves (**Figure 10**) revealed an interesting dichotomy between theoretical discriminative capability and practical classification performance. While normal tissue achieved a promising AUC of 0.913, the actual classification decisions proved less reliable, suggesting potential issues with decision boundary optimization. Additionally, Precision-Recall curves (**Figure 11**) for normal tissue classification highlights fundamental challenges in feature discrimination.

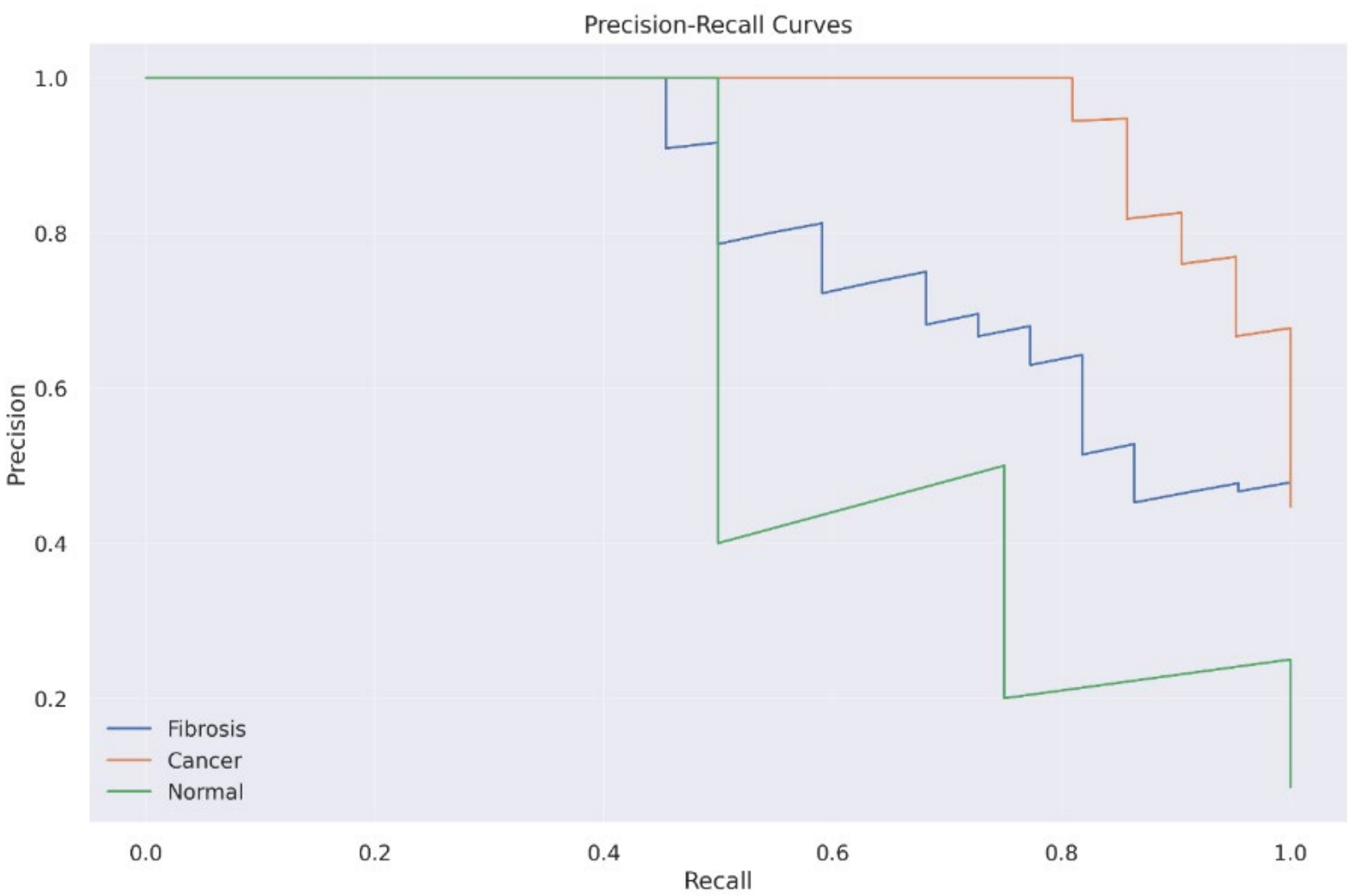

**Figure 11**. Precision-Recall curves highlighting performance trade-offs across different operating points. The rapid precision degradation for normal tissue classification suggests fundamental challenges in feature discrimination.

### 3.3 Feature Learning Analysis

To better understand the behavior of the model, we performed an in-depth analysis of the learned feature representations. This analysis provided several important insights into how the network interprets and differentiates between various tissue types.

In the context of cancer detection, the model demonstrated strong performance, indicating that it successfully learned to identify distinctive patterns associated with malignant tissue. The high specificity achieved by the model suggests it effectively distinguishes cancerous samples from benign ones, reducing the risk of false positives. Moreover, the model's consistent performance across different decision thresholds indicates robust feature extraction and classification capability, reinforcing its reliability in varied clinical scenarios.

However, classification of normal tissue posed greater challenges. The model exhibited difficulty in capturing the subtle characteristics of normal tissue, which often lack the prominent features present in cancerous samples. Although the receiver operating characteristic (ROC) AUC remained high, the corresponding low precision implies that the model struggles to place an optimal decision boundary for normal tissue. This discrepancy points to potential overfitting to specific patterns in the normal tissue subset, limiting its generalization ability.

Analysis of the learned feature space further illustrated these trends. Features extracted from cancerous tissues formed clearly separable clusters, underscoring the network's ability to isolate malignancy-related signals. In contrast, features from normal and fibrotic tissues showed considerable overlap, suggesting that the network had difficulty learning discriminative boundaries between these two classes. This overlap highlights the need for improved feature learning strategies, particularly in enhancing the separability of benign tissue classes.

### 3.4 Binary Classification Performance

#### 3.4.1 Final Model Architecture

Following our experimentation of multi-class tissue differentiation, we changed our approach to focus on the fundamental clinical objective: cancer detection. This strategic simplification led us to develop a specialized binary classification model based on the ResNet-18 architecture. Our final implementation used a fully trainable network, allowing complete adaptation to the specific characteristics of pancreatic tissue imaging, while incorporating moderate dropout (0.1) and a hidden layer dimension of 512 units to maintain robust generalization.

The training was carefully crafted to address the challenges identified in our earlier experiments. We implemented a conservative learning strategy with a base rate of $1 \times 10^{-4}$ and weight decay of 0.01, complemented by label smoothing and mixup augmentation. This combination proved essential in preventing overfitting while maintaining the model's ability to capture subtle tissue characteristics. Gradient clipping at 1.0 further enhanced training stability, particularly important given our relatively small batch size of 16.

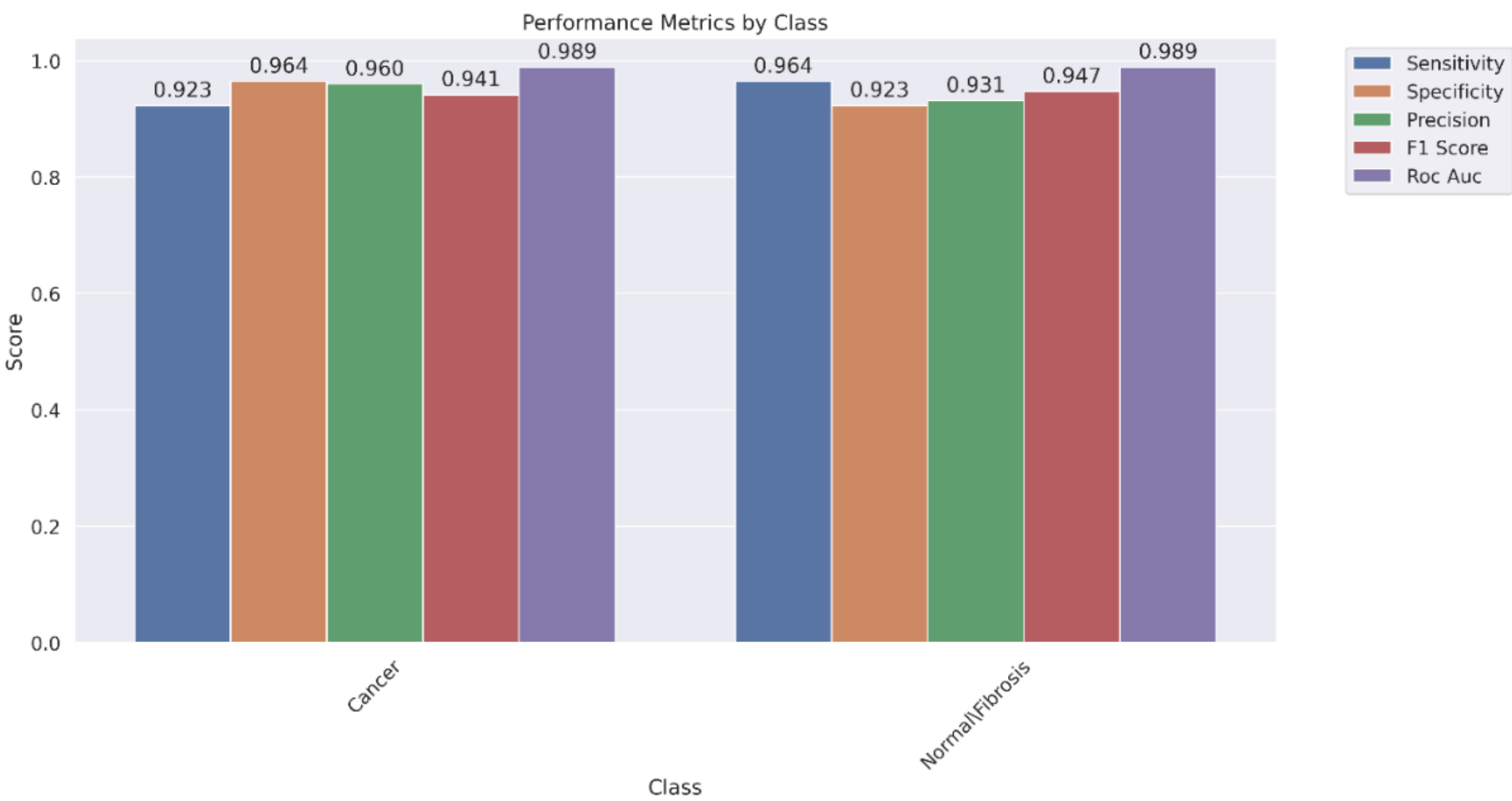

**Figure 12**. Performance metrics demonstrating exceptional classification capabilities across all measures. Both cancer and normal/fibrosis categories achieved ROC AUC values of 0.989, indicating near-perfect discrimination.

### 3.4.2 Performance Analysis

The binary classification model achieved outstanding performance across all key metrics, with both sensitivity and specificity reaching 0.964 (**Figure 12**). Particularly noteworthy is the precision of 0.960 for cancer detection, indicating extremely reliable positive predictions. The F1 score of 0.941 further demonstrates the model's exceptional balance between precision and recall. Most impressively, the model achieved an ROC AUC of 0.989 for both classes, approaching perfect classification.

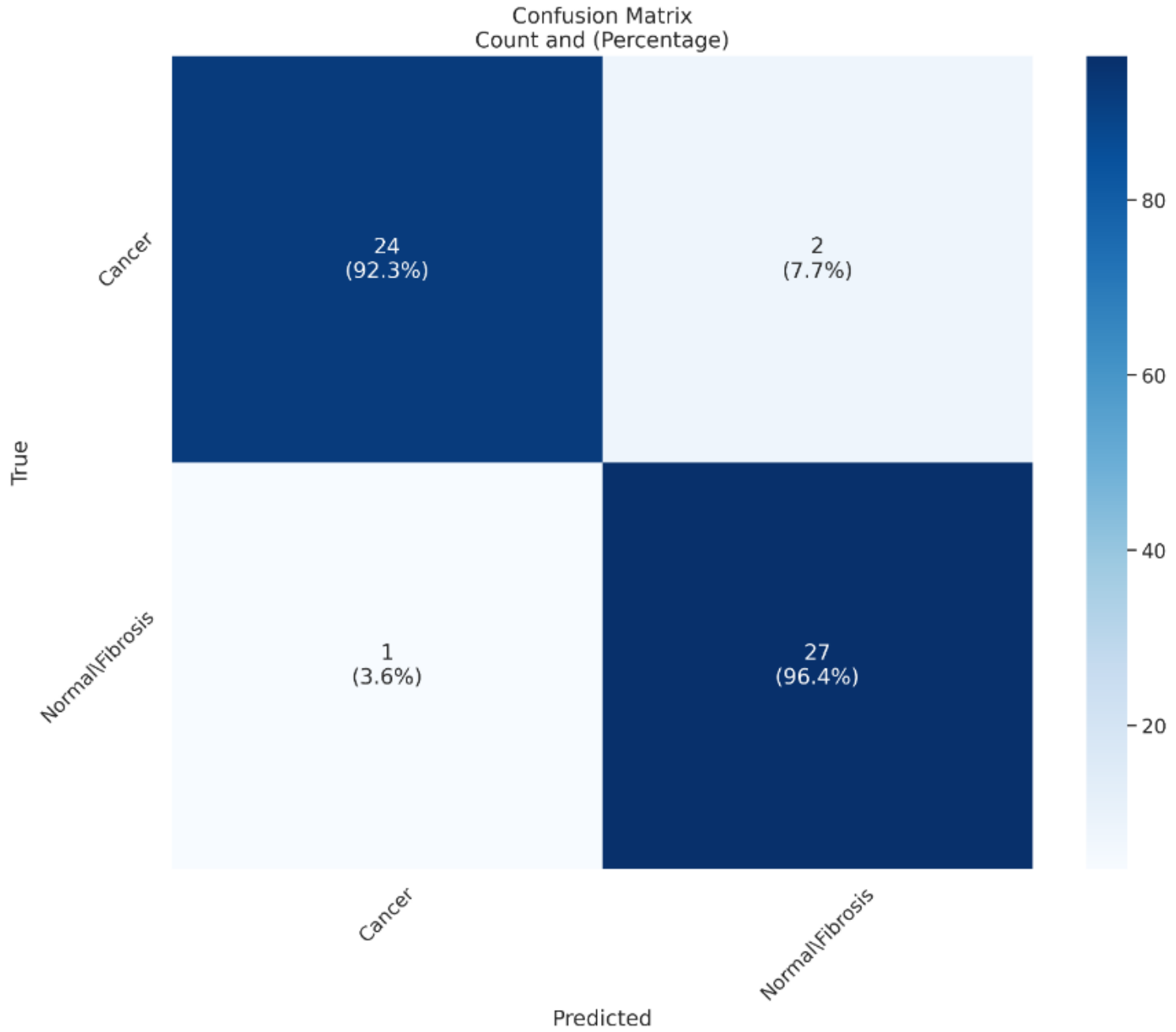

**Figure 13**. Confusion matrix revealing highly accurate classification with minimal errors. The model achieved 92.3% accuracy for cancer detection (24/26 cases) and 96.4% accuracy for normal/fibrotic tissue identification (27/28 cases).

The confusion matrix (**Figure 13**) reveals particularly strong performance, with the model correctly identifying 24 out of 26 cancer cases (92.3%) and 27 out of 28 normal/fibrotic cases (96.4%). This high accuracy in both categories is especially significant given the challenging nature of pancreatic tissue analysis and the critical importance of minimizing both false positives and false negatives.

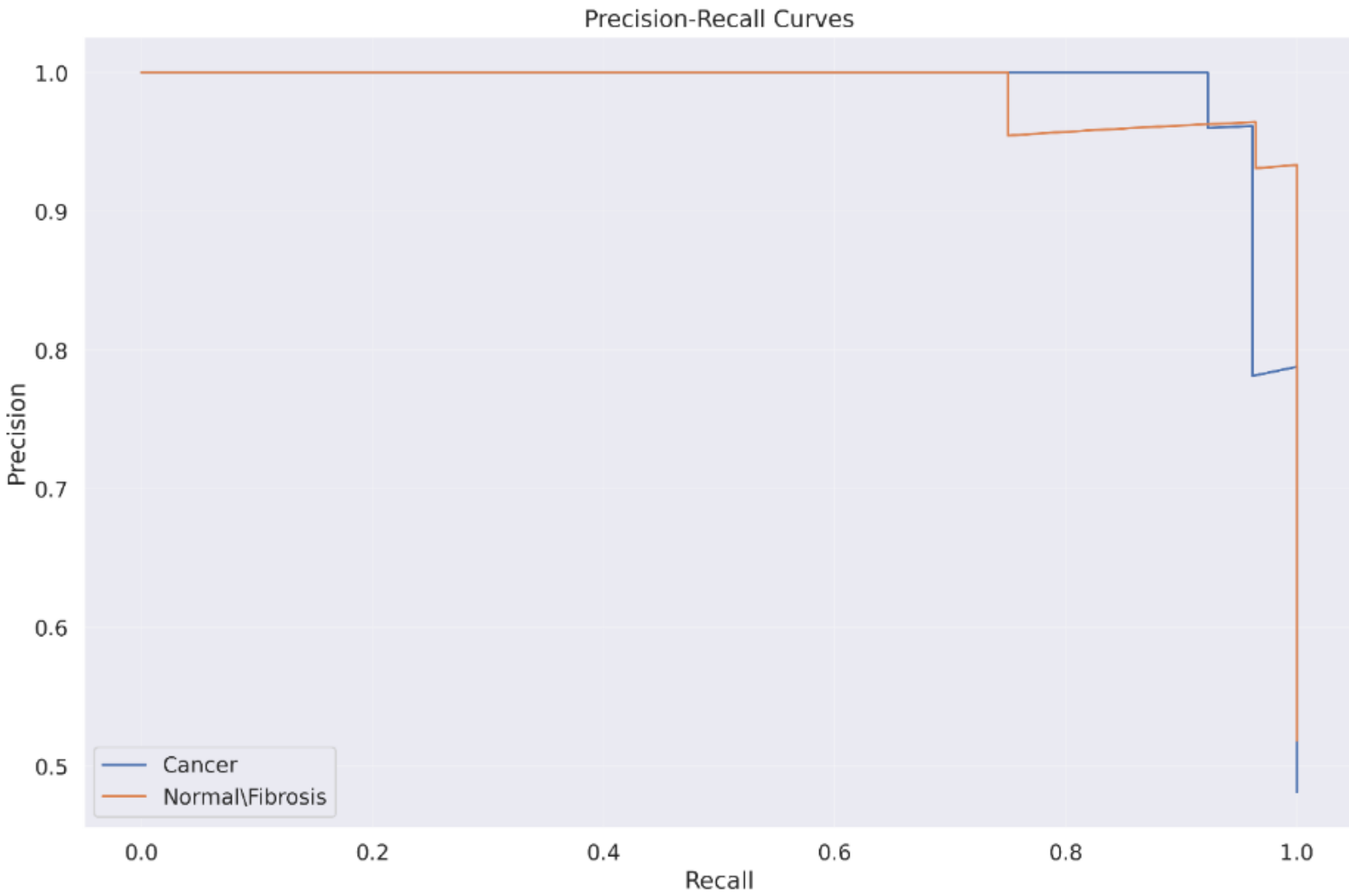

**Figure 14**. Precision-Recall curves demonstrating exceptional stability across different classification thresholds. Both cancer and normal/fibrosis classes maintain high precision even at increased recall levels.

The precision-recall (**Figure 14**) characteristics provide additional evidence of the model's robust performance. Most notably, the curves maintain remarkably high precision (above 0.9) across a broad range of recall values, suggesting extremely stable performance across different operating points. This stability is particularly valuable in clinical settings, where threshold adjustment may be necessary to optimize for specific sensitivity or specificity requirements.

The exceptional performance characteristics of our final model suggest potential for clinical application. The balanced accuracy of 0.964 for both sensitivity and specificity addresses the critical requirements in cancer detection - minimizing both missed malignancies and false positives. The outstanding ROC AUC (**Figure 15**) of 0.989 demonstrates nearly perfect discriminative ability, providing strong confidence in the model's decision-making capabilities across different operating conditions.

These results represent a significant advance in automated PDAC detection. The model's exceptional performance across all metrics, combined with its stable behavior across different operating points, indicates strong potential for clinical impact as an assistive technology for pathologists. This would be particularly valuable in early-stage screening scenarios, where the model's high sensitivity and specificity could significantly improve detection rates while minimizing unnecessary follow-up procedures.

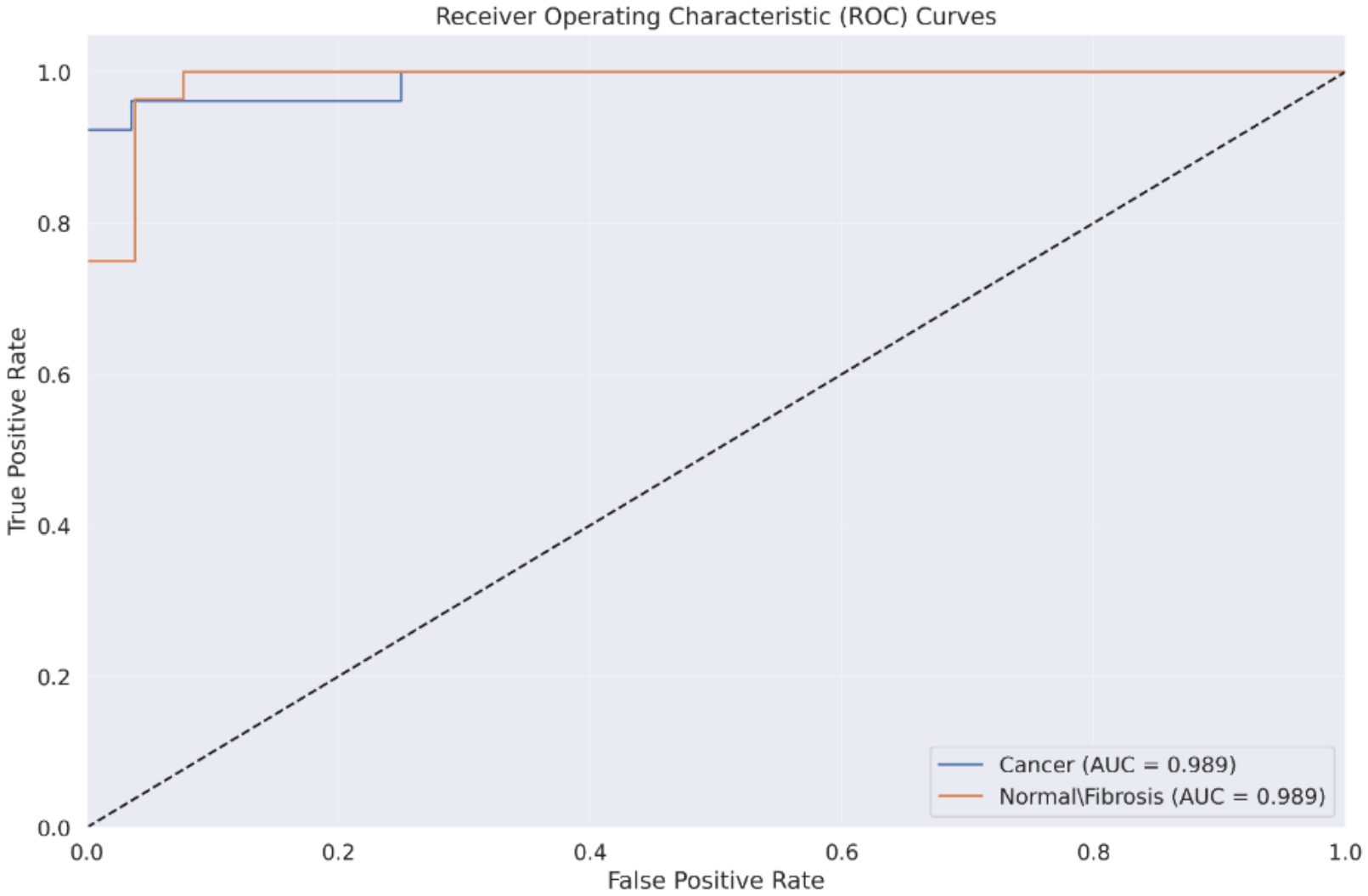

**Figure 15**. ROC curves showing exceptional discriminative ability with AUC of 0.989 for both classes, representing near-perfect separation between cancer and non-cancer cases.

### 3.5 Model Interpretability Analysis

To provide insight into the model's decision-making process, we used Gradient-weighted Class Activation Mapping (Grad-CAM) visualization. This technique highlights the regions of input images that most strongly influence the model's classifications, offering valuable transparency into the features driving the model's decisions.

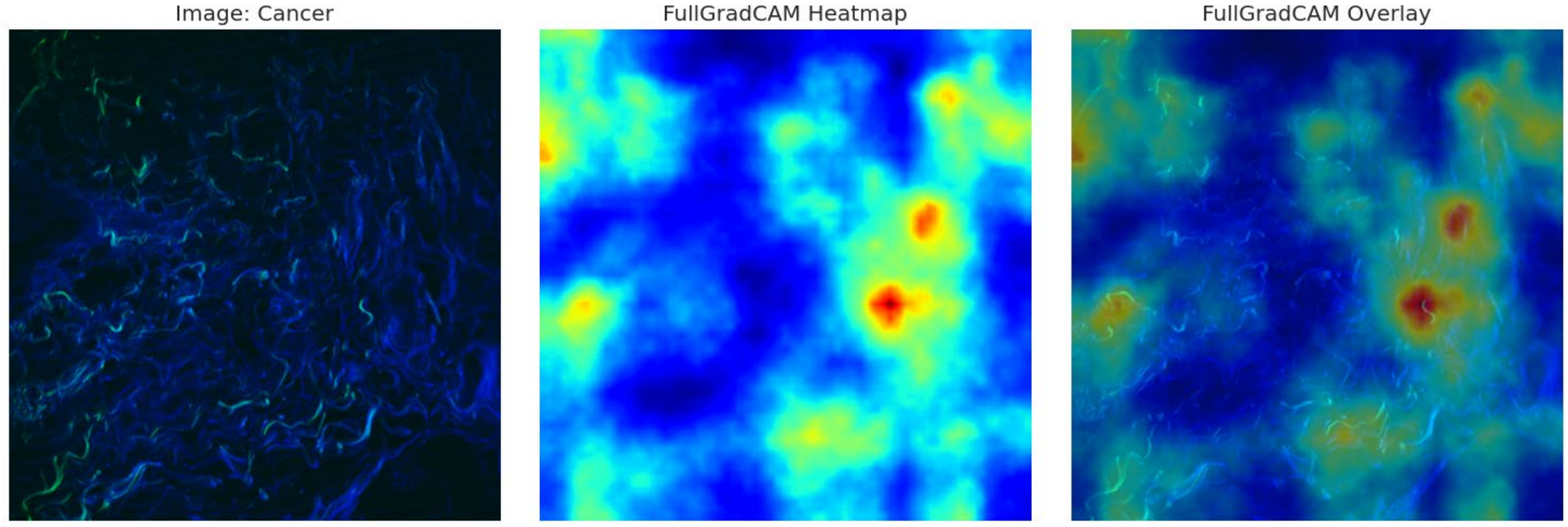

**Figure 16**. Grad-CAM visualization of cancer tissue classification. Left: Original dualmodality image. Center: Activation heatmap showing regions of interest. Right: Overlay showing the model's attention patterns when identifying cancerous tissue. The intense activations (red regions) highlight areas the model finds most significant for classification.

Analysis of the cancer tissue Grad-CAM visualizations (**Figure 16**) reveals distinct attention patterns in the model's classification process. The activation heatmaps show that the model consistently focuses on specific regions and structural patterns within the images when identifying cancerous tissue. These patterns appear to align with meaningful image features rather than random or artifactual elements.

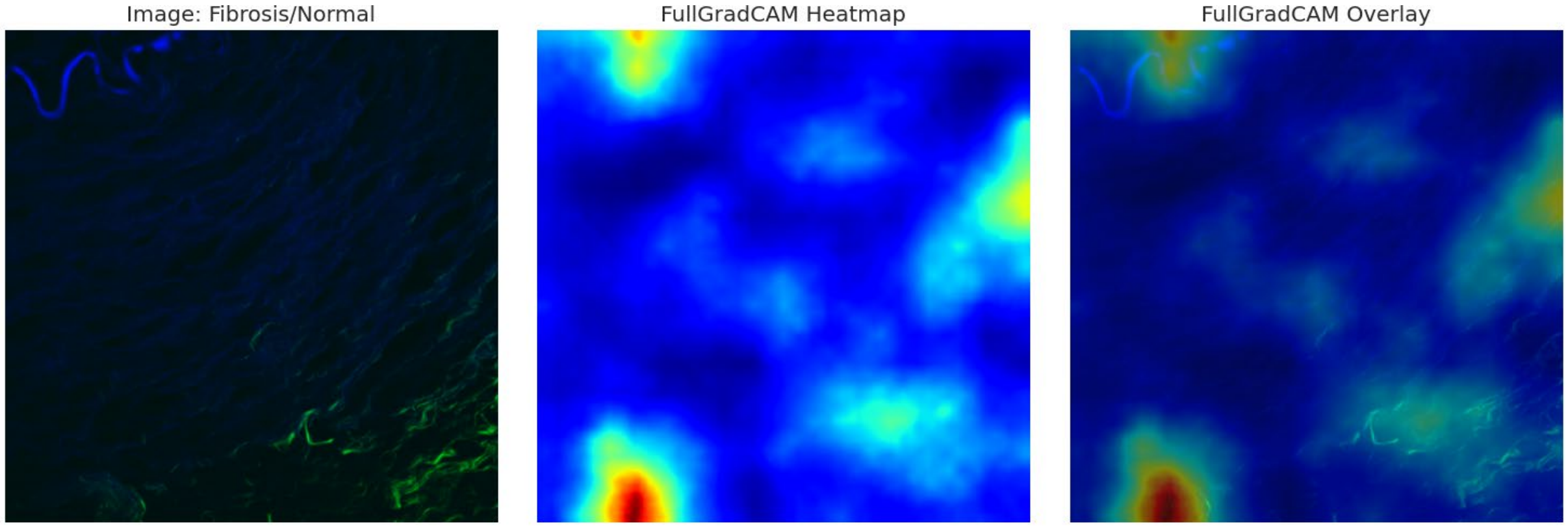

**Figure 17**. Grad-CAM visualization of normal/fibrotic tissue classification. Left: Original dual-modality image. Center: Activation heatmap. Right: Overlay demonstrating different attention patterns when classifying non-cancerous tissue.

When analyzing normal and fibrotic tissue (**Figure 17**), the model demonstrates notably different activation patterns compared to cancer cases. The heatmaps reveal that the model attends to distinct structural features when classifying non-cancerous tissue, suggesting it has learned to differentiate between tissue types based on underlying image characteristics.

The analysis of Grad-CAM visualizations for non-cancerous tissue revealed several notable characteristics that enhance our understanding of the model's internal reasoning. The activation patterns differed substantially between cancerous and non-cancerous classifications, indicating that the model employs distinct recognition strategies based on the underlying tissue type. In non-cancerous samples, the model consistently focused on specific structural elements within the images, suggesting it has learned to identify differentiating features that extend beyond superficial image characteristics. These heatmaps support the notion that the model is not merely memorizing patterns but systematically learning to extract meaningful representations relevant to tissue classification.

A deeper exploration of the Grad-CAM outputs provided critical insights into feature identification and pattern recognition. The model exhibited a consistent focus on particular regions within the tissue images, with reproducible attention patterns observed across similar cases. This consistency indicates the presence of learned discriminative features that the model uses to distinguish between tissue types. Moreover, systematic differences in activation patterns between classes were evident, underscoring the model's ability to apply distinct recognition strategies depending on whether the input was normal, fibrotic, or cancerous tissue. The attention maps showed reliable emphasis on biologically relevant characteristics, reinforcing confidence in the model's interpretability.

These results also speak directly to the model's decision transparency. The Grad-CAM visualizations offered clear, interpretable representations of the areas that influenced the model's classifications, thereby enabling validation of its decision-making process. This transparency is crucial in medical AI applications, where clinicians must trust and understand the system's recommendations. By visualizing attention regions and aligning them with known pathological features, the model enhances both the explainability and accountability of AI-assisted diagnoses.

Overall, the visualization findings support the model's potential as an interpretable diagnostic tool. The ability to rapidly identify regions of interest, coupled with a transparent decision-making process, makes the system well-suited for integration into existing diagnostic workflows. These strengths not only increase clinician trust but also promote the responsible deployment of deep learning in high-stakes clinical environments.

## 4 Limitations and Future Work

While our results demonstrate significant progress in automated PDAC detection, several areas warrant further investigation:

1. Dataset Expansion: Although our current dataset provided sufficient data for developing a robust binary classifier, expanding the dataset would enable exploration of more sophisticated architectures and potentially improve performance on edge cases.

2. Multi-modal Integration: Future work could explore the integration of additional imaging modalities or clinical metadata to enhance diagnostic accuracy. Particularly promising would be the incorporation of temporal data to track tissue changes over time.

3. Prospective Validation: Clinical implementation would benefit from prospective studies validating the model's performance across different patient populations and varying imaging conditions.

4. Model Optimization: While our current implementation achieves high accuracy, further optimization could focus on reducing computational requirements for real-time analysis and improving model efficiency for resource-constrained environments.

5. Extended Applications: The methodologies developed in this research could be adapted for other types of cancer detection, particularly those where early diagnosis significantly impacts patient outcomes.

## 5 Conclusion

This research has established a robust deep learning framework for early detection of pancreatic ductal adenocarcinoma through imaging analysis. Through systematic experimentation and optimization, we developed a highly accurate binary classification system that achieves 96.4% sensitivity and specificity in distinguishing between cancerous and non-cancerous tissue. The final architecture, based on a modified ResNet-18 model, demonstrates exceptional performance with an ROC AUC of 0.989, a decent metric for classification accuracy.

Our progression revealed multiple key points for applying deep learning to limited medical imaging datasets. The initial exploration of various architectures, including modern Vision Transformers and traditional CNNs, demonstrated that models often underperform simpler architectures when working with constrained dataset sizes. The success of our frozen backbone approach, combined with carefully tuned data augmentation and balanced training strategies, provides a valuable template for future medical imaging applications facing similar data limitations.

The developed framework offers significant potential for improving PDAC detection in clinical settings. The model's high sensitivity ensures reliable detection of malignant tissue, while its strong specificity minimizes false positives that could lead to unnecessary procedures. Particularly noteworthy is the model's ability to maintain consistent performance across different operating points, as demonstrated by the precision-recall characteristics, suggesting robust applicability across varying clinical requirements.

The integration of Grad-CAM visualization provides an additional layer of clinical utility by highlighting specific regions of interest that influence the model's decisions. This interpretability feature aligns with pathologists' diagnostic processes, potentially accelerating the identification of suspicious areas while maintaining transparency in the AI-assisted decision-making process. The visualization results demonstrate that the model bases its classifications on biologically relevant features, particularly focusing on collagen organization patterns that correspond to established pathological markers.

The success of our approach in developing a highly accurate classification system with limited training data has implications beyond PDAC detection. The framework we established, particularly the combination of transfer learning, balanced training, and comprehensive validation strategies, provides a valuable template for developing AI diagnostic tools in other medical domains facing similar data constraints.

Furthermore, our results demonstrate the potential for AI to augment rather than replace clinical expertise. The high performance of our relatively simple architecture, combined with visualization tools, suggests that focusing on clinical integration and may be more valuable than pursuing increasingly complex architectures.

**Acknowledgment:** The authors gratefully acknowledge the Institute of Medicine at the University of Maine, Orono, for supporting this research. We also acknowledge Dr. Mark Jones, MD, from MaineHealth for segmenting the regions of interest on the initial H&E slides, Dr. Karissa Tilbury for valuable discussions and insights regarding the data acquisition and interpretation, and Gerren Welch, a student researcher, for conducting the majority of the image acquisition.

**Conflict of Interest:** The authors declare that they have no conflict of interest.

**Data Availability Statement:** The code and integration scripts supporting this research are available in a public GitHub repository at https://github.com/deslo-research/pdac_ml.

**References**

(1) Sarantis, P.; Koustas, E.; Papadimitropoulou, A.; Papavassiliou, A. G.; Karamouzis, M. V. Pancreatic Ductal Adenocarcinoma: Treatment Hurdles, Tumor Microenvironment and Immunotherapy. *World J Gastrointest Oncol* **2020**, *12* (2), 173–181. https://doi.org/10.4251/wjgo.v12.i2.173.

(2) Renjifo-Correa, M. E.; Fanni, S. C.; Bustamante-Cristancho, L. A.; Cuibari, M. E.; Aghakhanyan, G.; Faggioni, L.; Neri, E.; Cioni, D. Diagnostic Accuracy of Radiomics in the Early Detection of Pancreatic Cancer: A Systematic Review and Qualitative Assessment Using the Methodological Radiomics Score (METRICS). *Cancers* **2025**, *17* (5), 803. https://doi.org/10.3390/cancers17050803.

(3) Liao, W.-C. Early Detection of Pancreatic Cancer: Opportunities Provided by Cancer-Induced Paraneoplastic Phenomena and Artificial Intelligence. *Journal of Cancer Research and Practice* **2023**, *10* (4), 129. https://doi.org/10.4103/ejcrp.eJCRP-D-23-00002.

(4) Yao, L.; Zhang, Z.; Keles, E.; Yazici, C.; Tirkes, T.; Bagci, U. A Review of Deep Learning and Radiomics Approaches for Pancreatic Cancer Diagnosis from Medical Imaging. *Current Opinion in Gastroenterology* **2023**, *39* (5), 436. https://doi.org/10.1097/MOG.0000000000000966.

(5) Moglia, A.; Cavicchioli, M.; Mainardi, L.; Cerveri, P. Deep Learning for Pancreas Segmentation on Computed Tomography: A Systematic Review. *Artif Intell Rev* **2025**, *58* (8), 220. https://doi.org/10.1007/s10462-024-11050-4.

(6) Hayashi, H.; Uemura, N.; Matsumura, K.; Zhao, L.; Sato, H.; Shiraishi, Y.; Yamashita, Y.; Baba, H. Recent Advances in Artificial Intelligence for Pancreatic Ductal Adenocarcinoma. *World Journal of Gastroenterology* **2021**, *27* (43), 7480–7496. https://doi.org/10.3748/wjg.v27.i43.7480.

(7) Anghel, C.; Grasu, M. C.; Anghel, D. A.; Rusu-Munteanu, G.-I.; Dumitru, R. L.; Lupescu, I. G. Pancreatic Adenocarcinoma: Imaging Modalities and the Role of Artificial Intelligence in Analyzing CT and MRI Images. *Diagnostics* **2024**, *14* (4), 438. https://doi.org/10.3390/diagnostics14040438.

(8) Alves, N.; Schuurmans, M.; Litjens, G.; Bosma, J. S.; Hermans, J.; Huisman, H. Fully Automatic Deep Learning Framework for Pancreatic Ductal Adenocarcinoma Detection on Computed Tomography. arXiv December 2, 2021. https://doi.org/10.48550/arXiv.2111.15409.
(9) Liu, H.; Gao, R.; Grbic, S. AI-Assisted Early Detection of Pancreatic Ductal Adenocarcinoma on Contrast-Enhanced CT. arXiv March 14, 2025. https://doi.org/10.48550/arXiv.2503.10068.
(10) Faur, A. C.; Lazar, D. C.; Ghenciu, L. A. Artificial Intelligence as a Noninvasive Tool for Pancreatic Cancer Prediction and Diagnosis. *World J Gastroenterol* **2023**, *29* (12), 1811–1823. https://doi.org/10.3748/wjg.v29.i12.1811.
(11) Marti-Bonmati, L.; Cerdá-Alberich, L.; Pérez-Girbés, A.; Díaz Beveridge, R.; Montalvá Orón, E.; Pérez Rojas, J.; Alberich-Bayarri, A. Pancreatic Cancer, Radiomics and Artificial Intelligence. *British Journal of Radiology* **2022**, *95* (1137), 20220072. https://doi.org/10.1259/bjr.20220072.
(12) Sijithra, P.; Santhi, N.; Ramasamy, N. A Review Study on Early Detection of Pancreatic Ductal Adenocarcinoma Using Artificial Intelligence Assisted Diagnostic Methods. *European Journal of Radiology* **2023**, *166*, 110972. https://doi.org/10.1016/j.ejrad.2023.110972.
(13) Li, H.; Lin, K.; Reichert, M.; Xu, L.; Braren, R.; Fu, D.; Schmid, R.; Li, J.; Menze, B.; Shi, K. Differential Diagnosis for Pancreatic Cysts in CT Scans Using Densely-Connected Convolutional Networks. arXiv June 19, 2018. https://doi.org/10.48550/arXiv.1806.01023.
(14) Ramaekers, M.; Viviers, C. G. A.; Janssen, B. V.; Hellström, T. A. E.; Ewals, L.; van der Wulp, K.; Nederend, J.; Jacobs, I.; Pluyter, J. R.; Mavroeidis, D.; van der Sommen, F.; Besselink, M. G.; Luyer, M. D. P. Computer-Aided Detection for Pancreatic Cancer Diagnosis: Radiological Challenges and Future Directions. *Journal of Clinical Medicine* **2023**, *12* (13), 4209. https://doi.org/10.3390/jcm12134209.
(15) Chen, C.; Qin, C.; Qiu, H.; Tarroni, G.; Duan, J.; Bai, W.; Rueckert, D. Deep Learning for Cardiac Image Segmentation: A Review. *Front. Cardiovasc. Med.* **2020**, *7*. https://doi.org/10.3389/fcvm.2020.00025.
(16) Enriquez, J. S.; Chu, Y.; Pudakalakatti, S.; Hsieh, K. L.; Salmon, D.; Dutta, P.; Millward, N. Z.; Lurie, E.; Millward, S.; McAllister, F.; Maitra, A.; Sen, S.; Killary, A.; Zhang, J.; Jiang, X.; Bhattacharya, P. K.; Shams, S. Hyperpolarized Magnetic Resonance and Artificial Intelligence: Frontiers of Imaging in Pancreatic Cancer. *JMIR Medical Informatics* **2021**, *9* (6), e26601. https://doi.org/10.2196/26601.
(17) Tilbury, K. B.; Hocker, J. D.; Wen, B. L.; Sandbo, N.; Singh, V.; Campagnola, P. J. Second Harmonic Generation Microscopy Analysis of Extracellular Matrix Changes in Human Idiopathic Pulmonary Fibrosis. *JBO* **2014**, *19* (8), 086014. https://doi.org/10.1117/1.JBO.19.8.086014.
(18) Hamilton, J.; Breggia, A.; Fitzgerald, T. L.; Jones, M. A.; Brooks, P. C.; Tilbury, K.; Khalil, A. Multiscale Anisotropy Analysis of Second-Harmonic Generation Collagen Imaging of Human Pancreatic Cancer. *Frontiers in Oncology* **2022**, *12*.
(19) Wen, B.; Campbell, K. R.; Tilbury, K.; Nadiarnykh, O.; Brewer, M. A.; Patankar, M.; Singh, V.; Eliceiri, K. W.; Campagnola, P. J. 3D Texture Analysis for Classification of Second Harmonic Generation Images of Human Ovarian Cancer. *Sci Rep* **2016**, *6* (1), 35734. https://doi.org/10.1038/srep35734.
(20) Tilbury, K.; Han, X.; Brooks, P. C.; Khalil, A. Multiscale Anisotropy Analysis of Second-Harmonic Generation Collagen Imaging of Mouse Skin. *JBO* **2021**, *26* (6), 065002. https://doi.org/10.1117/1.JBO.26.6.065002.
(21) Drifka, C. R.; Loeffler, A. G.; Mathewson, K.; Keikhosravi, A.; Eickhoff, J. C.; Liu, Y.; Weber, S. M.; Kao, W. J.; Eliceiri, K. W. Highly Aligned Stromal Collagen Is a Negative Prognostic Factor Following Pancreatic Ductal Adenocarcinoma Resection. *Oncotarget* **2016**, *7* (46), 76197–76213. https://doi.org/10.18632/oncotarget.12772.
(22) Gant, K. L.; Patankar, M. S.; Campagnola, P. J. A Perspective Review: Analyzing Collagen Alterations in Ovarian Cancer by High-Resolution Optical Microscopy. *Cancers* **2024**, *16* (8), 1560. https://doi.org/10.3390/cancers16081560.

(23) Kučikas, V.; Werner, M. P.; Schmitz-Rode, T.; Louradour, F.; van Zandvoort, M. A. M. J. Two-Photon Endoscopy: State of the Art and Perspectives. *Mol Imaging Biol* **2023**, *25* (1), 3–17. https://doi.org/10.1007/s11307-021-01665-2.

(24) Zhang, Y.; Akins, M. L.; Murari, K.; Xi, J.; Li, M.-J.; Luby-Phelps, K.; Mahendroo, M.; Li, X. A Compact Fiber-Optic SHG Scanning Endomicroscope and Its Application to Visualize Cervical Remodeling during Pregnancy. *Proceedings of the National Academy of Sciences* **2012**, *109* (32), 12878–12883. https://doi.org/10.1073/pnas.1121495109.

(25) Montague, J.; Young, L.; Shir, H.; Sawyer, T. W.; Routh, J.; Nfonsam, V. N.; Barton, J. K. Feasibility of Non-Imaging, Random-Sampling Second Harmonic Generation Measurements to Distinguish Colon Cancer. *BIOS* **2024**, *1* (3), 035001. https://doi.org/10.1117/1.BIOS.1.3.035001.

(26) Nannan, L.; Gsell, W.; Belderbos, S.; Gallet, C.; Wouters, J.; Brassart-Pasco, S.; Himmelreich, U.; Brassart, B. A Multimodal Imaging Study to Highlight Elastin-Derived Peptide pro-Tumoral Effect in a Pancreatic Xenograft Model. *Br J Cancer* **2023**, *128* (11), 2000–2012. https://doi.org/10.1038/s41416-023-02242-w.

(27) He, K.; Zhang, X.; Ren, S.; Sun, J. Deep Residual Learning for Image Recognition. In *Proceedings of the IEEE conference on computer vision and pattern recognition*; 2016; pp 770–778.

(28) Dosovitskiy, A.; Beyer, L.; Kolesnikov, A.; Weissenborn, D.; Zhai, X.; Unterthiner, T.; Dehghani, M.; Minderer, M.; Heigold, G.; Gelly, S. An Image Is Worth 16x16 Words: Transformers for Image Recognition at Scale. *arXiv preprint arXiv:2010.11929* **2020**.

(29) Gorriz, J. M.; Clemente, R. M.; Segovia, F.; Ramirez, J.; Ortiz, A.; Suckling, J. Is K-Fold Cross Validation the Best Model Selection Method for Machine Learning? arXiv November 8, 2024. https://doi.org/10.48550/arXiv.2401.16407.